\documentclass[11pt]{article}
\pdfoutput=1
\usepackage{multirow}
\usepackage{authblk}
\usepackage{comment}
\newcommand{\done}{\cellcolor{magenta}}  

\usepackage{pdfpages}
\usepackage{xcolor,colortbl}
\usepackage{adjustbox}
\usepackage{amsmath}
\usepackage{algorithm}
\usepackage{algpseudocode}
\usepackage{array}
\usepackage{mathtools}
\usepackage{subfig}
\usepackage{lineno}
\usepackage[bookmarks=false]{hyperref}
\usepackage{setspace}
\usepackage{epstopdf}
\begin{document}


\title{An End-to-End Breast Tumour Classification Model Using Context-Based Patch Modelling- A BiLSTM Approach for Image Classification}

\author[1,*]{Suvidha Tripathi}
\author[1]{Satish Kumar Singh}
\author[2]{Hwee Kuan Lee}
\affil[1]{Department of Information Technology, Indian Institute of Information Technology Allahabad, Devghat, Jhalwa, Prayagraj-211015, India}
\affil[2]{School of Computing, National University of Singapore,13 Computing Drive, 117417, Singapore, 
\newline Bioinformatics Institute, A*STAR, 30 Biopolis Street, 138671, Singapore,
\newline Image and Pervasive Access Lab(IPAL), CNRS UMI 2955, 1 Fusionopolis Way, 138632, Singapore,
\newline Singapore Eye Research Institute, 20 College Road, 169856, Singapore}
\affil[*]{Corresponding author: Suvidha Tripathi, suvitri24@gmail.com}
\date{}

\maketitle
\begin{abstract}
Researchers working on computational analysis of Whole Slide Images (WSIs) in histopathology have primarily resorted to patch-based modelling due to large resolution of each WSI. The large resolution makes WSIs infeasible to be fed directly into the machine learning models due to computational constraints. However, due to patch-based analysis, most of the current methods fail to exploit the underlying spatial relationship among the patches. In our work, we have tried to integrate this relationship along with feature-based correlation among the extracted patches from the particular tumorous region. For the given task of classification, we have used BiLSTMs to model both forward and backward contextual relationship. RNN based models eliminate the limitation of sequence size by allowing the modelling of variable size images within a deep learning model. We have also incorporated the effect of spatial continuity by exploring different scanning techniques used to sample patches. To establish the efficiency of our approach, we trained and tested our model on two datasets, microscopy images and WSI tumour regions. After comparing with contemporary literature we achieved the better performance with accuracy of 90\% for microscopy image dataset. For WSI tumour region dataset, we compared the classification results with deep learning networks such as ResNet, DenseNet, and InceptionV3 using maximum voting technique. We achieved the highest performance accuracy of 84\%. We found out that BiLSTMs with CNN features have performed much better in modelling patches into an end-to-end Image classification network. Additionally, the variable dimensions of WSI tumour regions were used for classification without the need for resizing. This suggests that our method is independent of tumour image size and can process large dimensional images without losing the resolution details.
\end{abstract}




\section{Introduction}
Image based computational pathology has developed into an ever-evolving field for computer vision researchers. New methods are being introduced frequently in this field for natural everyday scenes, face recognition, video analysis and other forms of biometrics. Despite that, the rate of development of medical image CAD algorithms for enhancing their diagnostic performance could not mirror the rate of development of new natural scenes analysis algorithms. It may have been due to the highly heterogeneous nature of cancer cells, which increases the complexity of the task at hand. In context with breast cancer, extensive research based on oncogenic pathways and tumor cell metabolism, and based on chemotherapeutic observations, it has been realized by pathologists that the disease is quite unpredictable \cite{leong2011changing}. 
\begin{figure}[htp]
\centering
\includegraphics[width=\textwidth ,height=3.5in]{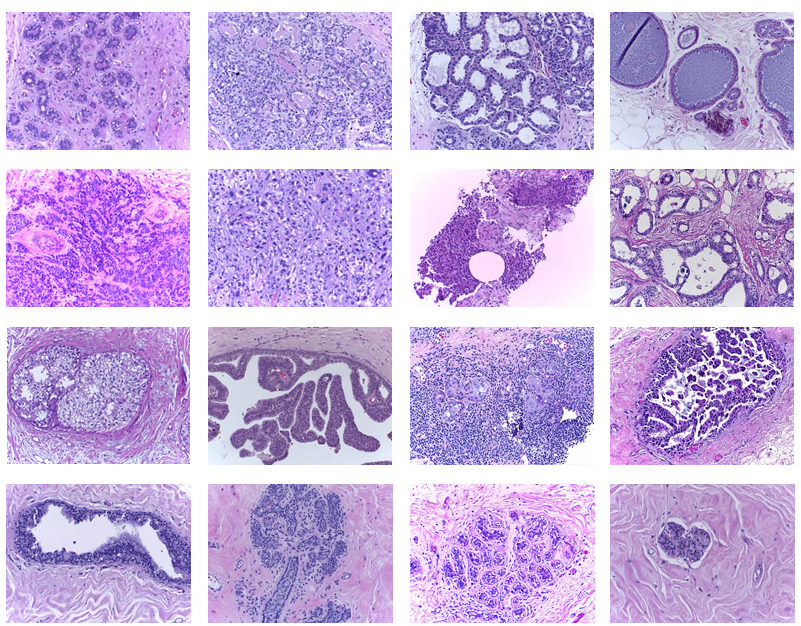}
\caption{Microscopy BACH data samples. First row to fourth: Benign Tumours, Invasive Carcinoma, \textit{In situ} Carcinoma, and Normal}
\label{micro}
\end{figure} 
 
\par Hence, there is a pressing need for the development of computer vision algorithms that are particularly advanced for diagnostic and prognostic evaluation of digitized biopsy images. Until then, there is a progressive adaptation of currently available state of the art methods for cancer detection, segmentation, and classification. The importance of precise prognosis in this field requires the differentiation of digitized samples into two, three, or more classes. In our work, we have four classes, Normal, Benign, \textit{In situ} carcinoma, and Invasive Carcinoma. The process of classification of breast samples by the pathologist help in a more accurate understanding of the disease and consequently help in the directed treatment of patients. The manual process is; however, quite a time consuming and requires an expert's knowledge due to the underlying complexity of the nature of images. The efforts to automate such non-trivial problem requires expert intervention to verify the diagnosis made by the CAD process. Besides that, the feasibility of implementation of such algorithms poses preliminary challenges. For instance, high resolution of gigapixel Whole Slide Images could not be processed by any current state of the art algorithms due to their large size. A large amount of information present in one patient slide makes the task more challenging concerning space and efficiency. Therefore, for practical problem solving, we need to either build new systems that could address such challenges or find a workaround of our problems that could be feasibly addressed by available systems.  One such workaround is dividing WSI into patches of the size that could be easily fed into the algorithm. However, this leads to loss of overall structure of the tumor and various other sub-structures present in the slide. The spatial continuity of the patches also becomes hard to incorporate within a deep end-to-end model. The task becomes more non-trivial in case of a 4-class  problem rather than a 2-class problem where the structures need to be segregated between two widely spaced classes. As the number of classes or segregation increases, the space between classes reduces. 
\par Considering all these issues, we chose our model such that the gigapixel size of WSI could be harnessed in a way without losing the structure of the overall suspected region. The spatial relationship between the patches of the same region could be modeled end-to-end without the need of building a separate algorithm to infuse the context of the previous patch in the sequence of the patches which together makes an entire tumor region. 
\par One such state of the art model which is well known for preserving the contextual relationship, is Recurrent Neural network, commonly known as RNN. RNNs have been used by the computer vision community to process sequences such as texts and videos. We acknowledged its efficiency and formulated our problem around the strength of the RNNs, which is processing sequences of patches from the same region and eventually classify the input sequence as one of the four classes. We classified image regions as a whole using BiLSTMs. BiLSTMs are a known version of RNNs for modeling textual and video sequences. They have been widely used for activity recognition in videos and have proved their niche in modeling future contextual information due to their bi-directional architecture. Our method could serve its purpose in clinical diagnosis by assisting pathologists for labeling suspected regions automatically.
Our main contribution is summarised in following points:
\begin{enumerate}
\item According to our knowledge, this is the first study that includes the use of contextual information among the patches from the same region using BiLSTMs for classification of tumors.
\item Our method is robust to the size of the tumour regions as it can take both very huge dimensions WSI and microscopy regions. In this study, the range of tumour regions vary between 17290 to 236 pixels across height and 20570 to 195 pixels across the width. 
\item The study did not alter the size of the tumors for deep modeling and classify variable size tumor regions by processing them as a sequence of features.
\item This work proposed end-to-end network for patch to image classification unlike previous literatures that use stage wise networks to first classify patches and then aggregate classification results of patches into image labels \cite{hou2016patch,nazeri2018two,mahbod2018breast,
wang2019rmdl,roy2019patch,huang2018improving, shaban2019context, araujo2017classification}.

\item This is a shallow network that do not require heavy training to train hundreds of layers as in ResNet and GoogleNet. 

\item We also experimented with patch scanning methods to verify that a particular scanning technique that deploy maximal connectivity between patches is better than randomly extracting patches from the image. 
 
\end{enumerate}  

\label{intro}
\section{Related Work}
The application of RNN based architectures such as LSTM \cite{hochreiter1997long} and BiLSTMs \cite{schuster1997bidirectional,graves2005bidirectional} on series data classification such as texts and time series has been a very common methodology. Researchers have recently started combining CNNs and LSTMs for image captioning  \cite{johnson2016densecap, karpathy2015deep, vinyals2015show} or multi-label image classification \cite{zhang2018multilabel, wang2016cnn, wei2014cnn, guo2018cnn} as well. The idea of using RNN based models for image classification stemmed from the fact that objects in an image are often, though not always, related to each other in some way. Images, although, are not sequential data but carry some latent semantic dependencies which can be modeled as a sequence of occurrences of certain objects present in the image that overall define the global image description. These deep LSTMs based models have, however, are not sufficiently explored on high-resolution medical data. With high dimensional images in case of WSIs, the tumor regions when divided into patches can act as a sequential data that have some contextual dependency with each other. Modeling this contextual information among patches is a crucial step to perform slide level classification. \par There have been studies in Whole Slide Image level analysis that have drawn contextual and spatial relationship among patches using their novel methods. For instance, authors in \cite{huang2018improving} proposed a deep spatial fusion network to predict image-wise label from patch-wise probability maps. They evaluated their network performance on two datasets BIC \cite{Bioimaging2015} and BACH \cite{bach2018} and used heavy augmentation due to the small volume of images. Their network was not end-to-end and required heavy data pre-processing steps to enhance the performance. They used microscopy images to test their model, which have dense class properties. Whereas, in the case of Whole Slide Image annotations, the tumour class like Invasive carcinoma could be spread across the gigapixel image and the parts of the annotation may look like normal. Therefore, with WSIs, the parts of the annotation when broken into patches, may not give the reliable label. Hence, such methods should be tested on such datasets as well for better clinical significance. The method in \cite{shaban2019context} exploits the spatial context between patches extracted from high resolution histopathological images for grading of colorectal cancer histology images. The authors propose a two staged framework consisting of two stacked CNNs. The first CNN called as LR-CNN learns the representations of the patches and aggregates the learned features from each patch in the same spatial dimension as the original image ($M\times N$). So, in other words LR-CNN converts a high resolution image into high dimensional feature map. The next stage consists of context aware blocks called as RA-CNN that takes feature representation cube as input to learn the spatial relationship between patches to make a context-aware prediction.  The authors explored different network architectures for context-aware learning. 	The strategy solves a huge challenge of missing contextual information in patch-based classifiers. The robustness of the method also lies in the fact that the use of pre-trained architectures to extract features reduces the time and effort to train large models. However, the authors did not test their method on WSIs which pose a challenge of multi-resolution feature learning and very large size. In case of WSIs the feature cube could be as large as high resolution images and then its processing in a deep learning network could become infeasible.
\par To address the problem of multi-resolution analysis, majority of the previous works of literature have used patch-level analysis which requires breaking up of structures and hence global level features are lost. But, due to multi-resolution data, it is in fact left as an only choice to process such images.  All the methods using WSI datasets discussed above have done the same for developing their models. Studies like \cite{hou2016patch,nazeri2018two,mahbod2018breast,wang2019rmdl,roy2019patch}  have performed patch-based modelling of histopathology slides or microscopy images to perform image-wise classification using methods like probability fusion and majority voting. The authors in \cite{wang2019rmdl} developed a two-stage processing pipeline for classifying WSIs of gastric cancer. The first stage- discriminative instance selection selected the most informative patches on the basis of probability maps generated by a localization network. The second stage performed the image level prediction. The authors proposed a novel  recalibrated multi-instance deep learning network (RMDL) with the purpose of aggregating both local and global features of each instance via a modified local-global feature fusion module. RMDL framework presented an effective way to aggregate patches for final image level prediction by exploiting the interrelationship of the patch features and overcame the drawbacks of direct patch aggregation. The method is however limited in its approach as it is confined to same scale context and do not address the spatial relationship between the instances.  
\par The authors in \cite{spanhol2016breast} studied the applicability of deep learning architectures in identifying the breast cancer malignant tumours from benign tumours. The different sets of experiments were designed to train the CNN with different strategies that allow both high and low resolution images as input.
\par In \cite{bayramoglu2016deep} two CNN architectures have been used to identify breast cancer tumour and the magnification of the image. Single Task CNN classifies the benign and malignant tumour. Whereas, multi-task CNN has two output branches which takes multi-resolution image patches as input and produces two classification – between malignant and benign and between four classes of magnification. 
\par Similarly, Araujo et al. \cite{araujo2017classification} first proposed a patch-wise classification and then combined the patch probabilities to perform image-wise classification. They used their custom CNN model to perform patch-wise classification and achieved 66.7\% accuracy. Then the majority voting scheme was used among the classified patches to predict the overall image label. This method was also not end-to-end and required extensive CNN training and experiments to decide optimal hyper-parameters for their proposed model. They also did not consider spatial context among the patches to build a relationship between same image patches which may have proved crucial performance enhancer.
\par  All these methods although solve the challenge of multi-resolution analysis by patch-level aggregation of classification results, suffer from lack of spatial context and continuity relationship among patches. Moreover, due to the inherent limitation of state-of-the-art deep learning models which takes only a fixed size input, the previous works of literature had to sometimes perform heavy resizing to conform to the size of network input. Therefore, CNN + RNN based model could be the perfect replacement of such models since they could provide both spatial and contextual modelling, strategic region extraction method without the limitation of resizing along with the end-to-end compact model to process high-resolution Whole Slide and microscopy images. 
\par Few of the recent pieces of literatures have used such type of CNN + RNN models for the analysis of histopathological data. For instance, the paper \cite{qaiser2019learning} explores the application of deep reinforcement learning in predicting the diagnostically relevant regions and their HER2 scores in breast immunohistochemical (IHC) Whole Slide Images. For the given task, the authors proposed context module and a CNN-LSTM end-to-end model. The model intelligently views the WSI as the environment and the CNN-LSTM acts as a decision maker or the agent. 
Their model successfully mimics the histopathological expert analysis that first looks coarsely at ROIs at low resolution and then predict the scores of diagnostically relevant regions. Their model also incorporates multi-resolution analysis by combining features of the same region at multiple resolution for better predictive performance. The main advantage given by their model is that one need not look at all the regions of a WSI to predict the outcome and instead could focus on small number of regions without sacrificing the performance of the model. Similarly, \cite{ren2018differentiation} and \cite{bychkov2018deep} have also used the combination of CNN-LSTM for disease outcome prediction. The authors \cite{ren2018differentiation} have used the genomic data (Pathway Scores PS) with disease recurrence extracted from gene expression signatures exhibited in prostate tumors with a Gleason 7 score to identify prognostic marker. They calculated the PS scores and combined them with deep learning model for the purpose of combining the prognostic markers with image biomarkers. The deep learning model used is CNN-LSTM end-to-end model that take WSI patches as input sequence. CNN finds the features which LSTM processes to output the final hazard ratios of recurrence of the disease. Thy compared their model performance with different image features (LBP, HOG, SURF, neurons) with pathway scores. The results shows higher hazard rations with CNN-LSTM + PS in comparison to other clinically relevant prognostic features used in the comparison. The model show a novel idea of combining genetic markers with image biomarkers using LSTM in their model in order to preserve the spatial and contextual relationship among patches. However, the model is not sufficiently validated with different datasets along with their choice of CNN model and choice of training parameters. The paper \cite{bychkov2018deep} predicts the five-year disease specific survival of patients diagnosed with colorectal cancer directly from digitized images of haematoxylin and eosin (H\&E) stained diagnostic tissue samples. The authors used a CNN-LSTM based model that takes TMA spots as input sequence into the model. The VGG16 architecture was used to extract patch features. The model claims the novelty of providing direct outcome prediction instead of doing intermediate analysis like classifying tissue samples. The proposed model by the authors used different scanning techniques to extract patches but claimed to have found no effect on the final prediction results. This claim is not properly validated in the study and is contradictory to what we found in our experimental analysis. The authors compared their model with traditional machine learning classifiers such as naïve-bayes, logistic regression, SVM. The lack of comparison with contemporary deep learning classifiers weakens the validation of the proposed method. All these methods using CNN-LSTM as their base model have shown the applicability of RNN based models in disease prognosis. Keeping the advantages in mind, we used the BiLSTMs, the Bidirectional LSTM to classify tumour regions in our work. The experimental observations on our dataset (Section \ref{experiments}) showed the advantage of using BiLSTMs over LSTMs in our model.   
\label{review}
\section{Methodology}
\subsection{\textbf{Overview}}
In medical images, patch level classification is often useful for detecting cancer in microscopy and WSI images. However, if the prediction needs to be made for a whole tumor or gland, the network model needs to be trained such that the whole tumor region could be classified without losing its structure, resolution, and spatial correlation. For building such model, we have first extracted annotated tumour regions from WSIs and performed rotation transformation on regions for rotation invariance. After pre-processing of WSI dataset, we divided both microscopy and WSI tumour samples into patches. The patches were acquired by following different scanning techniques. We further developed a BiLSTM network model that takes the patches acquired from large tumour regions in the form of sequences. Since, the patches were extracted in a continuous pattern, therefore, we were able to construct a sequential data fit for BiLSTM network. 
We extracted features from each patch in a sequence using GoogleNet (pre-trained on ImageNet). Accumulated features per region formed one sequence. The sequences were then passed through BiLSTM layers for classification into labels. At the test time, the test regions follow the same feature extraction and sequence formation procedure. The trained BiLSTM model then tests the sequence and give out the predicted label. In brief, the method follows the 5 steps: 1)extracting whole regions (Benign, Invasive, and \textit{In situ}), 2) extracting patches from each tumor region, 3) extracting features from each set, per patch, 4)forming a sequence out of each set, and 5) sequence processing and classification. 
\label{overview}
\subsection{\textbf{Preprocessing}}
\subsubsection{Region Extraction}
\label{re}
\begin{figure}[htbp]
\centering
\includegraphics[width=\textwidth ,height=4in]{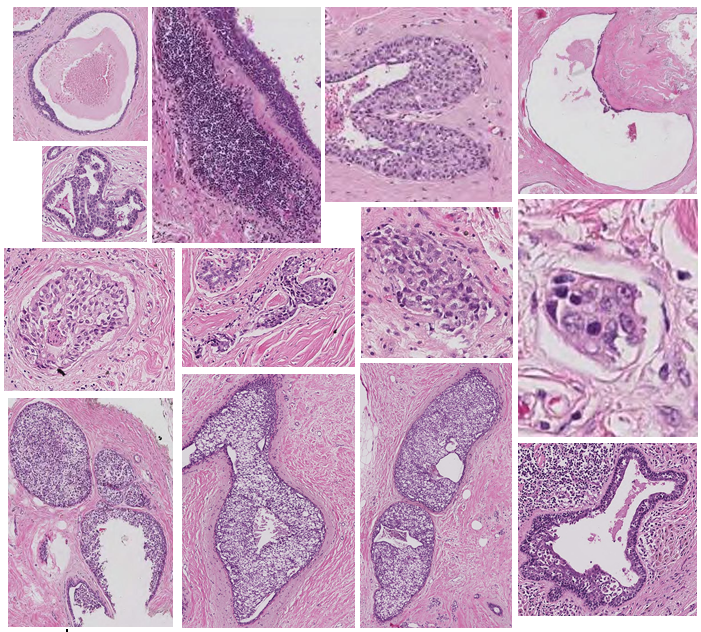}
\caption{WSI BACH data samples extracted from gigapixel slides. The variable size of each tumour region pose limitation in traditional deep learning framework. But, our model mitigate this limitation by allowing variable sequence size. First row to third: Benign Tumours, Invasive Carcinoma, \textit{In situ} Carcinoma. These regions can be seen having different dimensions but represent a single resolution level (level 0) from the WSI pyramid (Fig. \ref{WSIpyramid})}
\label{WSI}
\end{figure}

The histopathological breast cancer slide dataset used in our work contains ten annotated WSIs labeled into four major classes, Normal, Benign, \textit{In situ} carcinoma, Invasive carcinoma. The annotation of each WSI is recorded in XML files. Each XML file is divided into regions as annotated by pathologists in the corresponding WSI. The regions are then marked by drawing a rough boundary around the suspected region. The boundary is marked using slide annotation tools such as ASAP (Automated Slide Annotation Platform). Each pixel coordinate annotated by the pathologist is recorded in the XML file under a current region being annotated. The XML file also contains the region label, area of the region in pixels, region id, zoom ration, length of the region in microns, and area of the region in microns. Each annotated coordinate is represented in X, Y, and Z  axes values. From the available information, we calculated the maximum and minimum boundary coordinates to find out the location, height, and width of the labeled region.  
\par Since the tumour regions can be found in varying orientations depending upon the angle of acquisition of the particular WSI or microscopy image, the model should be robust to such changes. Therefore, to make the process more robust and rotation invariant, the obtained regions were rotated by following a unified method. To determined the angle of rotation for a particular region, the region mask was used to analyse the orientation of the region with respect to the vertical axis. The angle of rotation was then calculated following the steps below:
\begin{enumerate}
\item Determine the major axis centroid of the region. 
\item Calculate the major axis angle ($M$) from the X-axis.
\item Calculate the angle of rotation $R= 90-M$
\item Rotate the region along the major axis centroid by the angle $R$.
\item Repeat steps 1 to 4 for both region and region mask
\item Calculate the bounding box coordinates of the rotated mask. 
\item Modify the obtained bounding box dimensions to the nearest multiple of 256. 
\item Crop rotated region around the modified bounding box coordinates. 
\end{enumerate}

\subsubsection{Scanning Methods for Patch Extraction}
\label{pe}
Some of the extracted regions had large pixel dimensions due to their high resolution, which required breaking regions into patches to enable the processing of the regions. The arbitrary dimensions of the sampled regions was also an issue for the deep network training since such networks require equal size images as input. Therefore, for the feasibility of the experiment, the regions were divided into patches of dimension $256 \times 256$. The particular patch size was chosen keeping in mind following points:
\begin{itemize}
\item The smaller patch size in the power of 2 is $128 \times 128$. This patch size contains less details than a $256\times 256$ patch. 
\item The larger patch size $512\times 512$ and more (in the powers of 2) although would contain more details and context, but will impose computational constraints like expensive computation resources and time. This scenario would not be feasible in hospital implementation and integration of the CAD methods. 
\item The pre-trained deep learning models like GoogLeNet, ResNet, DenseNet, InceptionV3 take fixed size input ranging from 200 to 300 pixels across their width and height. hence, taking smaller or larger patch sizes would demand heavy resizing resulting in loss of information and details. Therefore, $256\times 256$ patch size seemed appropriate for the proposed method. Many recent literatures like \textit{Wang et al.}, \textit{Chennaswamy et al.} in \cite{aresta2019bach} have resized their patches to $256\times 256$ and then resized them to $224 \times 224$ in order to process them with deep learning architectures like ResNet and DenseNet. 
\end{itemize}

\begin{figure} [htp] 
\centering
\vspace{-1in}
 \subfloat[short for lof][Scan\_1]{
   \includegraphics[width=0.45\textwidth, height=3in]{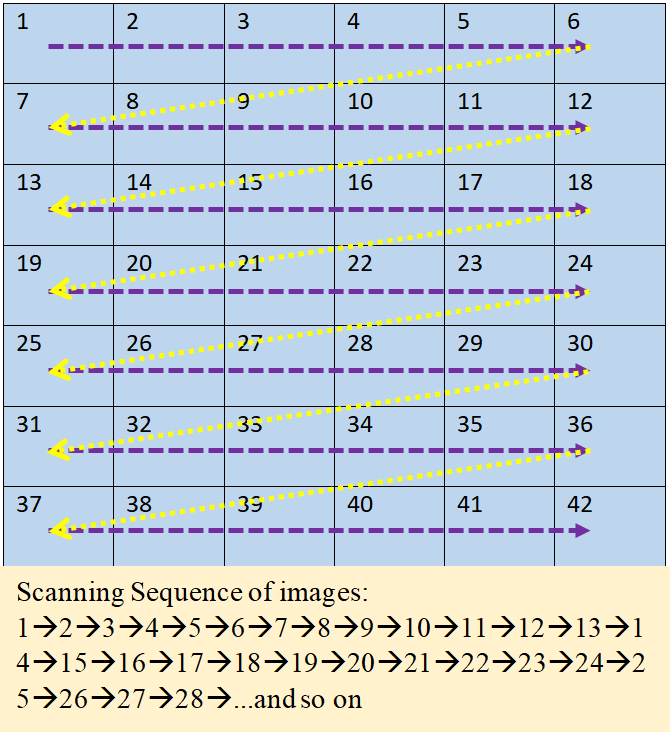}
   \label{subfig:fig1}
 }
 \subfloat[short for lof][Scan\_2]{
   \includegraphics[width=0.45\textwidth, height=3in]{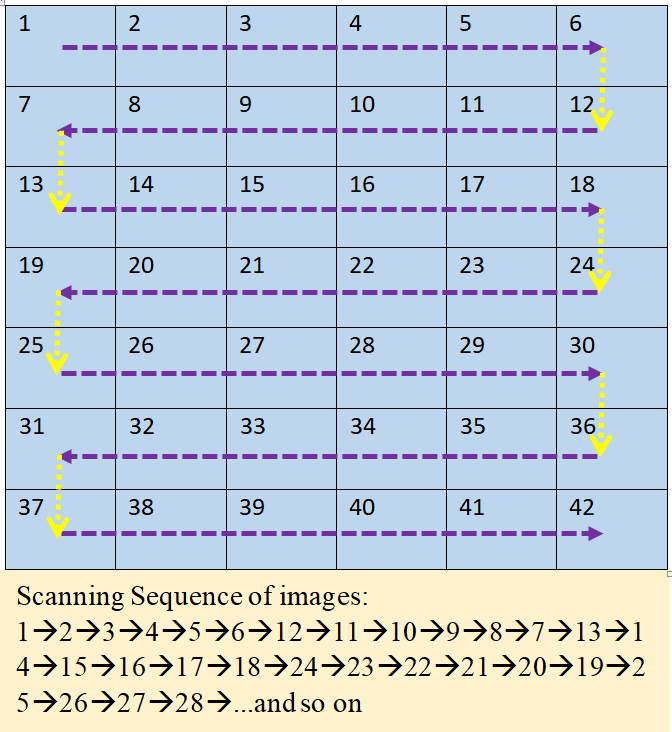}
   \label{subfig:fig2}
 }\\
 \subfloat[short for lof][Scan\_3]{
   \includegraphics[width=0.45\textwidth, height=3in]{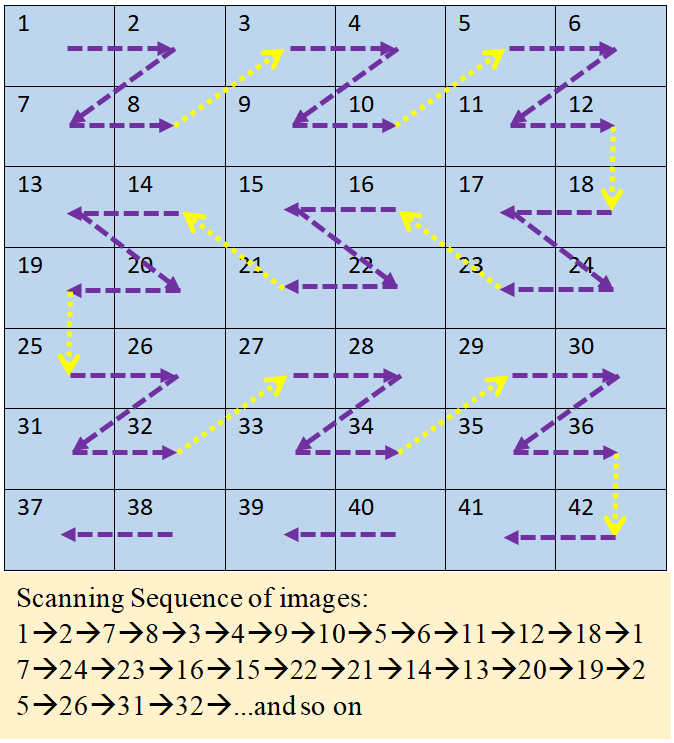}
   \label{subfig:fig3}
 }
\caption[short for lof]{The figure illustrates the different scanning methods that are used to extract patches from labeled WSI regions. The numbered blue blocks represents the patches in the WSI or Microscopy dataset. The dotted purple arrow shows the direction of scan and the dotted yellow arrow shows the transition from one pass of scan to another.}
\label{scan}
\end{figure}

To study and analyze the effect of different scanning techniques for sampling patches from regions, we tested three different scanning methods.  Fig. \ref{scan} shows the pictorial representation of these techniques. 
\par The first technique deploys most commonly used scanning method that moves the sliding window of desired patch dimensions from left to right across the width until the maximum width. The process is repeated across the height of the region. The window is non-overlapping, and at the extreme ends, if the expected height and/or width of the patch is greater than the remainder, we used symmetric padding to level the patch dimensions. For the convenience of the language, we addressed this scanning method as \textit{Scan\_1}. The process is illustrated in Fig. \ref{subfig:fig1}. 
\par The second scanning technique was thought as an attempt to arrange patches in sequence to bring as much continuity as possible. For any RNN method, where the sequence of data is the key to linking the context of the past and future with the present, we needed to derive sequential information from our tumor regions after they are sampled into patches. Our method is an effort to test the efficiency of RNN in case of image sequences. It scans patches starting from left to the right across the width in one iteration, and then the second iteration starts from the next row of non-overlapping pixels. It starts from right towards left, covering the width of the image. The process is repeated for subsequent rows until the entire region is exhausted. We named this scanning technique as \textit{Scan\_2}, shown in Fig. \ref{subfig:fig2}. 
\par The third scanning method was deployed to bring more correspondence between the neighboring patches. The patches were scanned as represented in Fig. \ref{subfig:fig3}. The set of four neighboring patches are scanned first, then the next adjacent batch, and onwards. When the row of non-overlapping pixels changes, the batches were scanned from right to left. The process was repeated until the region was covered across both dimensions. This technique is further referred to in the article as \textit{Scan\_3}.
The patches from each region were separated in the form of sets or folders. Each folder contained an arbitrary number of patches according to the dimensions of the particular region. The patches in each folder were labeled the same as the label of the region. 
\par The variable number of patches in each set does not limit the efficiency of our model. In fact, it allows huge dimensional tumour regions to be flexibly processed all at once in the form of a sequence without the need for heavy resizing. This flexibility pose as a strength of our model. 

\label{preprocessing}
\subsection{\textbf{CNN Feature Extraction}}
After patch sampling process using all three scanning techniques, each set of patches was passed through pre-trained GoogleNet architecture available in MATLAB 2019a for feature extraction step. The GoogleNet architecture was not fine-tuned on our datasets, and hence the hefty training process was not required in our work. The simple pre-trained weights of this architecture were used to extract deep features from the patches, and the sequence of features was constructed from each folder to be processed by BiLSTM layers. The complete process is further elaborated in the subsequent section. For comparison purposes during the experimental analysis, we also used ResNet101 and DenseNet201 pre-trained architectures to show the performance effect on the final classification output. 
\label{fe}
\subsection{\textbf{Tumour Region Classification}}
\label{rnn}

\begin{figure}[htp]
\centering
\vspace{-1in}
\includegraphics[width=\textwidth ,height=8in]{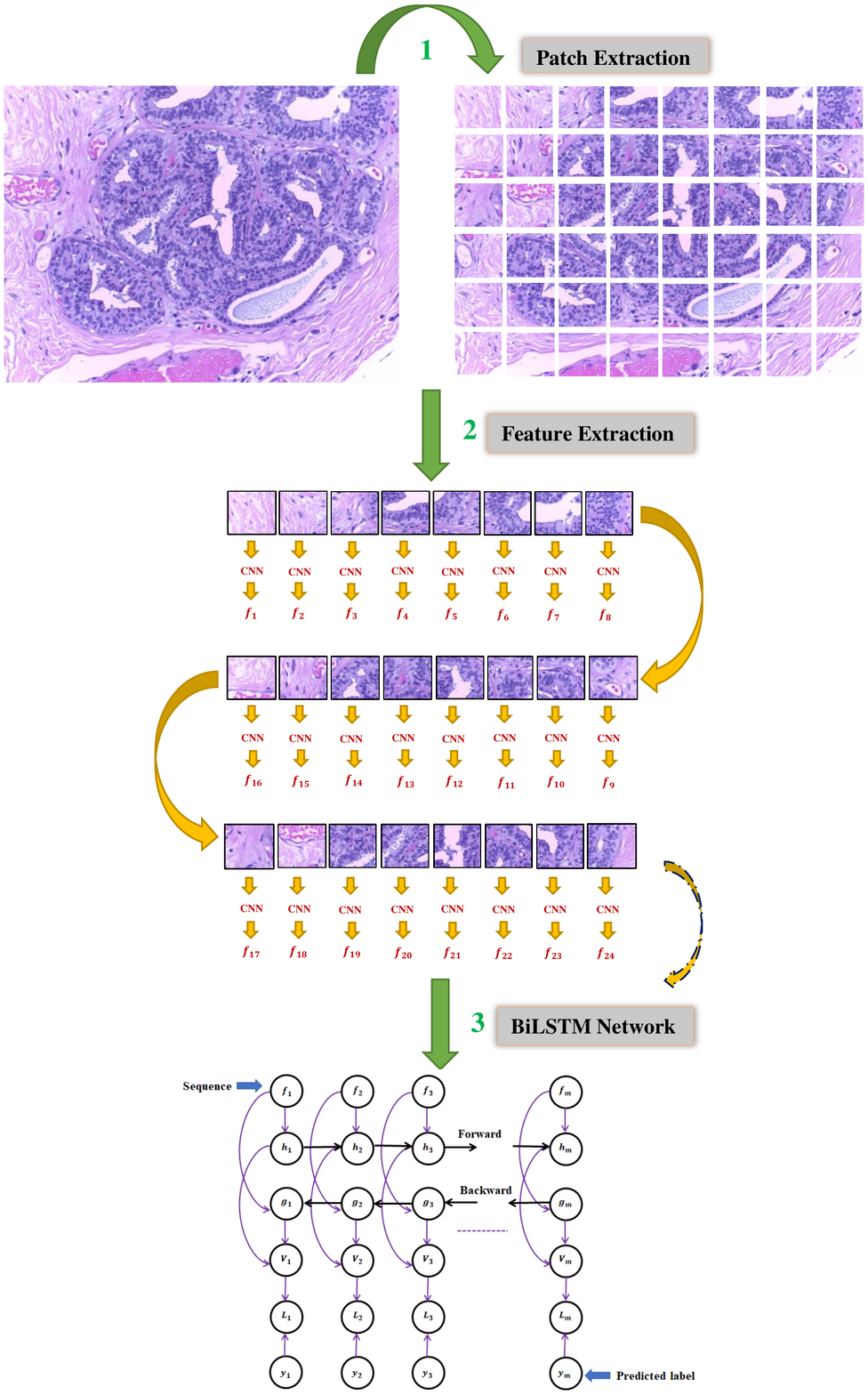}
\caption{Illustration of the whole process pipeline from patch extraction to computation of a BilSTM network. The steps are shown through green arrows. The first step is to extract patches, followed by CNN feature extraction and Sequence formation ($f_1,f_2,f_3,...$). The sequence of features is then used as a input to a BiLSTM network. BilSTM network is meant to learn to map input sequences $f$ to target sequences $y$. The $h$ recurrence propagates information forward in time (towards the right), while the $g$ recurrence propagates information backward in time (towards the left).}
\label{BiLSTM}
\end{figure} 

\subsubsection{\textbf{Patch Feature Sequence Formation}}
We used the GoogleNet pre-trained on ImageNet as a fixed feature extractor. The patches from each set are converted to sequences of feature vectors, where the feature vectors are the output of the activations function on the last pooling layer of the GoogleNet network ("pool-$7\times 7$\_s1"). We have used the pre-trained network because we did not have much training data to train a network from scratch, and there were no standard pre-trained weights publically available on similar medical data.
Each sequence is a D-by-m array, where D is the number of features (the output size of the pooling layer) and m is the number of patches in the region. Feature Dimension for one patch = 1024 X 1 (for GoogleNet features), feature dimension for $m$ patches in a region= $1024 \times m$; Labels for $m$ patches = label of the region. Let for patch 1 feature vector is $f_1$, patch 2 : $f_2$, patch 3 : $f_3, \ldots$, Patch $m$ : $f_m$, so the $nth$ sequence comprise : $f_1, f_2, f_3, f_4,\ldots, f_{m-1}, f_m$. Each labeled region in a WSI region forms a single sequence of patches. In other words, one region is converted into one sequence. We can then divide these sequences into training, testing, and validation sets. 
\label{sequence}
\subsubsection{\textbf{BiLSTM Training and Classification}}
BiLSTMs or Bi-directional Long Short Term Memory models are different from traditional LSTMs by capturing information from past as well as future. This type of network is feasible for applications where prediction depends on the whole input sequence. BiLSTMs combine two LSTMs where one LSTM take input sequence from start to end while the other LSTM takes input sequence from end to the first patch in the sequence. Figure \ref{BiLSTM} illustrates the BiLSTM model, with $h_{(t)}$ is the state of the sub-BiLSTM that moves forward through the ordered sequence and $g_{(t)}$ represents the state of the sub-BiLSTM that moves backward through the sequence where $t={1,2,3,\ldots,m}$ . The output unit $V_{(t)}$ is obtained by concatenating $h_{(t)}$ and $g_{(t)}$. $V_{(t)}$ is a representation that depends on both the past and the future of the sequence but is most sensitive to the current inputs.  
An output vector $V_{(t)}$ is calculated as 
\begin{equation}
V_{(t)}=f(h_{(t)}, g_{(t)})
\end{equation}
where function $f$ is used to combine the two output sequences. It can be a concatenating function, a summation function, an average function or a multiplication function. The following vector can represent the final output of a BiLSTM layer,
\begin{equation}
V_{m}=f(h_m, g_m)
\end{equation}
in which $V_m$, is the predicted sequence. Such a network where only the final output vector is sufficient to summarize a sequence is useful for predicting the label of the patch sequence. 
To Train this network, the cross-entropy loss function $L$ is used at the end to back-propagate information first through forward $h$ states and second through backward states  $g$. After forward and backward passes, the weights are updated. 
The sequence sets are passed through BiLSTM one at a time, and the predicted output tells the class of the sequence. We have used softmax classifier for our prediction.
The end-to-end network architecture is described in the Table \ref{network}
\begin{table*}[htbp]
\renewcommand{\arraystretch}{1.6}
\caption{End-to-end architecture of the Tumour classification network.}
\label{network}
\centering
\begin{adjustbox}{width=\textwidth}
\begin{tabular}{|c|c|c|c|c|}
\hline
\multirow{2}{*}{\textbf{Layer}}&\multirow{2}{*}{\textbf{Type}}&\multirow{2}{*}{\textbf{\shortstack{Input/Output\\ Dimensions}}}&\multirow{2}{*}{\textbf{Description}}\\
&&&\\
\hline
1&\shortstack{Sequence Input\\ Layer}&$224 \times 224 \times 3$&Enables sequence data input to a network \\
\hline
2&\shortstack{Sequence Folding\\ Layer}&\shortstack{Out:$224 \times 224 \times 3$\\ Minibatch: 1}& \shortstack{This layer enables processing of a batch of sequence \\input as a batch of images}\\
\hline
3-140&\shortstack{Convolution Layers \\(GoogleNet)}&\shortstack{Input: $224 \times 224 \times 3$\\ Output: $7 \times 7 \times 1024$}&\shortstack{All the middle layers of GoogleNet including convolution,\\ ReLU, Batch Normalization, Dropout, etc.} \\
\hline
141&\shortstack{Average Pooling Layer \\(pool5-$7\times 7$\_s1)}&$1 \times 1 \times 1024$&Average the the input feature dimension $(7 \times 7$) to ($1 \times 1$) \\
\hline
142&\shortstack{Sequence Unfolding \\Layer}&$1 \times 1 \times 1024$& \shortstack{Restores the sequential structure of the input sequence \\of images. Minibatch output of sequence folding layer is\\ connected to minibatch input of this layer.}\\
\hline
143&\shortstack{Flatten Layer}&$1024$&Reshapes the 3 dimensional feature vector to one dimension\\
\hline
144&\shortstack{BiLSTM Layer \\(2000 hidden units)}&$4000$&\shortstack{Enables learning bidirectional long term \\dependencies between sequence of patches from a region.\\Hidden units correspond to amount of information remembered \\between time steps or hidden states of BiLSTM}\\
\hline
145&\shortstack{Dropout Layer}&$4000$& \shortstack{Randomly sets input features to zero with a specified probability.\\ This is added to prevent network overfitting.}\\
\hline
146&\shortstack{Fully Connected Layer}&$3$&Multiplies the weight matrix and adds bias to the input features.\\
\hline
147&\shortstack{Softmax Layer}&$3$&Applies softmax function to the input\\
\hline
148&\shortstack{Classification Layer}&$-$&Computes the cross-entropy loss.\\
\hline
\end{tabular}   
\end{adjustbox}  
\end{table*}

\label{training}
\label{methodology}
\section{Setup and Results}
\subsection{\textbf{ICIAR 2018 BACH Dataset}}
The BACH (\textbf{B}re\textbf{A}st \textbf{C}ancer \textbf{H}istology) dataset was released by ICIAR 2018 conference organizers as a grand-challenge for classification and localization of tumors segregated by clinically relevant four classes. The dataset was released in two parts, microscopy and whole slide images. The microscopy image dataset contains 400 histology images, each with dimensions $2048 \times 1536$. The 400 microscopy images are subdivided into 4 classes, i.e., Normal, Benign, \textit{In situ} carcinoma, and Invasive carcinoma. The division is equal, having 100 images in each class, making it a balanced dataset (see Fig. \ref{micro}). According to the data released by conference organizers in \cite{aresta2019bach}, the images were annotated by two medical experts and those images were discarded where the two experts had any disagreements. The images are originally provided in \textit{.tiff} format and have three channels (RGB).  The organizers of the BACH 2018 challenge have also provided patient details in separate files for both microscopy and WSI data. They have labelled the patient ids from 1 till 39. The WSI images were extracted from patient 1 to 10 whereas Microscopy images were extracted from patient 11 to 39.  The excel files containing data of patient ids can be viewed and downloaded from the URL https://iciar2018-challenge.grand-challenge.org/Dataset/. Therefore, the two datasets are extracted from different patients and hence different samples from unique patients helped to validate the performance of the proposed method.
More detailed information about the microscopy dataset was provided in their article \cite{aresta2019bach} and the challenge website \cite{bach2018}. Since the organizers did not release the test dataset labels, hence we did not consider the test dataset (100 images) in our work. 
\par
\begin{figure}[htbp]
\centering
\includegraphics[width=3in ,height=2in]{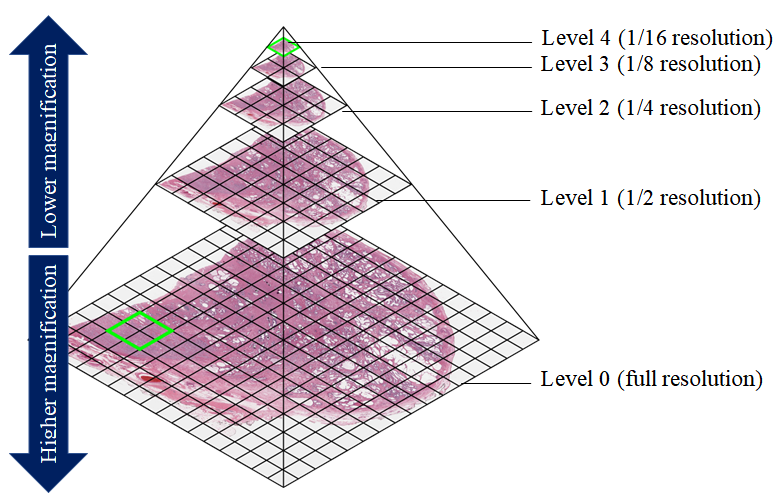}
\caption{WSI file pyramid structure}
\label{WSIpyramid}
\end{figure}

 The second dataset of Whole-slide images are high-resolution images of digitized sampled biopsy tissues. Each WSI contains more than one pathological labels (Benign, \textit{In situ} carcinoma, and Invasive carcinoma). All the unannotated regions are considered as normal. The challenge provided only 10 annotated WSIs and 20 unannotated WSIs for training. For testing, 10 more slides were released but, without labels. So, we had only 10 training WSI for both testing and training purposes.  From each WSI, after the region extraction step elaborated in Section \ref{re}, the distribution of labels is shown in Table \ref{table1}. The regions have different dimensions and were extracted at the highest resolution level. The understanding of resolution levels of WSI can be understood from Figure \ref{WSIpyramid}. Out of the 109 Invasive regions originally annotated by the pathologists, seven regions could not be read by the available computing resource due to their high dimension and memory constraints. Hence, we processed 102 Invasive regions in our work. The images were digitized in \textit{.svs} format and could only be accessed with ASAP or similar software. The organizers also provided the python code to read the annotation files. 
\begin{table}[htbp]
\renewcommand{\arraystretch}{2.3}
\caption{Distribution of the labels for the microscopy and WSI datasets}
\label{table1}
\centering
\begin{tabular}{|c|c|c|c|c|}
\hline
\textbf{Dataset}&\textbf{Benign}&\textit{\textbf{In situ}}&\textbf{Invasive}&\textbf{Normal}\\
\hline
\textbf{Microscopy}&100&100&100&100\\
\hline
\textbf{WSI}&57&109&60&-\\
\hline
\end{tabular}
\\ - denotes no annotated normal regions
\end{table}
\label{sec3.1} 
\subsection{\textbf{Dataset Preparation}}
The dataset obtained from the challenge had to be pre-processed for them to be feasibly used for input in the network. The high dimensional WSI regions for the purpose were broken into patches of size $256 \times 256$. The process is explained in Section \ref{pe}. The total accumulated patches from WSI regions were 16,934 for the three classes. The same process for patch extraction has been repeated for microscopy dataset where each image was of fixed dimension $2048 \times 1536$. For patch extraction step, Microscopy images were divided into a grid of $ 8 \times 6$ dimensions. We call it a grid of patches with each patch of $256 \times 256$ dimensions and total 48 patches were acquired from each histology microscopy image. From this dataset, total 19,200 patches were extracted. The accumulated patches from each region were then divided in the form of sets with variable patch numbers. 
\label{sec3.5.1}

\label{sec3.2}
\subsection{\textbf{Data Usage}}
The sequences of features formed after feature extraction process were divided into training, validation and testing sets in the ratio 0.7:0.15:0.15 during parameter selection experiments. After deciding the optimal parameters on the hold-out sets, we followed the 10 fold cross-validation test to verify our results. The distribution of the data during parameter selection experiments is shown in Table \ref{table2}.
\begin{table}[htbp]
\renewcommand{\arraystretch}{2.3}
\caption{Distribution of the data for the microscopy and WSI datasets into training, validation and testing sets (for parameter selection only (refer Section \ref{experiments}))}
\label{table2}
\centering
\begin{adjustbox}{width=3.5in}
\begin{tabular}{cccccc}
\hline
\textbf{Dataset}&&\textbf{Benign}&\textbf{Invasive}&\textit{\textbf{In situ}}&\textbf{Normal}\\
\hline
\multirow{3}{*}{\textbf{Microscopy}}&Train&66&75&76&63\\
\cline{2-6}
&Validation&18&11&12&20\\
\cline{2-6}
&Test&16&14&12&17\\
\hline
\multirow{3}{*}{\textbf{WSI}}&Train&34&74&45&-\\
\cline{2-6}
&Validation&11&13&9&-\\
\cline{2-6}
&Test&12&15&6&-\\
\hline
\end{tabular}
\end{adjustbox}\\
 - denotes no annotated normal regions
\end{table}
All the images extracted from microscopy and WSIs had dimensions $256 \times 256$ and they were used in their raw form without any color normalization and adjustment. 
\label{sec3.3}
\subsection{\textbf{Experiments}}
\label{experiments}
\par The pre-trained architecture for extracting deep features was selected through experiments. We tested the performance of end-to-end architecture using ResNet101 and DenseNet201 as feature extractors. The accuracy obtained with ResNet101, \textit{Scan\_2}, and WSI dataset was 63.64\% whereas with Microscopy dataset, we obtained the accuracy of 84.85\%. Similarly, the accuracy obtained with DenseNet201, \textit{Scan\_2}, and WSI dataset was 77.97\% whereas 71.19\% with Microscopy dataset. The choice of \textit{Scan\_2} and different hyper-parameters used to train the model with ResNet101 and DenseNet201 was validated in our next experiment.    
\par The second experiment comprise optimal hyper-parameter selection using heuristics and best scanning technique for histopathological images irrespective of the dataset. 
First, we experimented with three optimizing functions- Stochastic Gradient Descent with Momentum (SGDM), RMSprop, and ADAM. The specific hyper-parameters for each optimizer is summarized as:\\
SGDM- Momentum: $0.90$, \\RMSprop- SquaredGradientDecayFactor: $0.9900$, Epsilon: $1.0000e-08$,\\ ADAM- GradientDecayFactor: $0.9000$, SquaredGradientDecayFactor: $0.9900$, Epsilon: $1.0000e-08$.\\ These are the default parameter settings in MATLAB2019a and we used them as is.  We made the combinations of the chosen hyperparameters which were four dropout rates, three scanning techniques,  three optimization functions and two learning rates. The symbol '*' in-front of some of the accuracy values represents that the epochs were run keeping training option of validation patience at 5. The setting ensures that the training stops if the validation loss is larger than or equal to the previously recorded smallest loss for at most 5 times during the training or if the maximum number of epochs are exhausted, whichever is the earlier. In this setting, the number of epochs may or may not reach the maximum limit set at the start of the training. So, we performed all the 72 experiments with and without validation patience 5 for a maximum of 30 epochs. We have shown only the largest of the two accuracy values obtained from the two settings. The '*' indicates that the larger accuracy value is obtained with validation patience 5. So, in total, we conducted $4 \times 3\times 3\times 2 \times 2=144$ experiments for each dataset to select the optimal hyper-parameters. The experimental results are indicated in table \ref{table3} for 3-class classification of WSI tumour regions and table \ref{table4} for 4-class classification accuracy of Microscopy dataset. . Several deductions were made from the table \ref{table3}. Such as, across all the scanning methods, learning rate $10^{-4}$ performed better than learning rate $10^{-3}$. However, for second scanning method (\textit{Scan\_2}), both the learning rates performed closely with accuracy values falling in the range 80-88\%. Scanning method \textit{Scan\_3} followed closely in terms of frequency of accuracy values more than 80\%. When we kept the optimization function, learning rate and scanning method constant, the trend of accuracy across different drop-out rates signify the importance of tuning drop-out values during training custom models. With respect to optimization function and irrespective of the drop-out rates, the all-over analysis of the table \ref{table3} suggests that SGDM did not perform well in first two scanning methods (\textit{Scan\_1} and \textit{Scan\_2}) whereas, the gain in SGDM performance was observed in \textit{Scan\_3}. In case of ADAM, this optimization function could not enhance model's performance across all hyper-parameters except in \textit{Scan\_2} with learning rate $10^{-4}$. The optimization function RMSprop performed consistently better across scanning methods \textit{Scan\_2} and \textit{Scan\_3} irrespective of the learning rates and drop-out rates. The highest performance as can be seen in the table \ref{table3} for WSIs was given by \textit{Scan\_2}, RMSprop, 0.5 drop-out rate and $10^{-4}$ learning rate. The cell is highlighted in magenta. 

\begin{table}[htbp]
\renewcommand{\arraystretch}{1.6}
\caption{Accuracy (\%) obtained against different learning rates,  rates, optimizing functions and scanning techniques with respect to Whole Slide Images (3-classes).}
\label{table3}
\centering
\begin{adjustbox}{width=3.5in}
\begin{tabular}{|c|c|c|c|c|c|c|}
\hline
\multirow{2}{*}{\textbf{\shortstack{Scanning\\ Method}}}&\multirow{2}{*}{\textbf{\shortstack{Learning\\ Rate}}}&\multirow{2}{*}{\textbf{Optimizer}}&\multicolumn{4}{|c|}{\textbf{Dropout Rate}}\\
\cline{4-7}
&&&\textbf{0.4}&\textbf{0.5}&\textbf{0.6}&\textbf{0.7}\\
\hline
\multirow{6}{*}{\textbf{\textit{Scan\_1}}}&\multirow{3}{*}{\textbf{$10^{-4}$}}&SGDM&51.52&57.58&54.55&39.39\\
\cline{3-7}
&&RMSprop&72.73*&60.61*&66.67&63.64\\
\cline{3-7}
&&ADAM&69.70&69.70&66.67&63.64\\
\cline{2-7}
&\multirow{3}{*}{\textbf{$10^{-3}$}}&SGDM&72.73*&72.73&66.67*&66.67*\\
\cline{3-7}
&&RMSprop&60.61&66.67*&63.64*&57.58*\\
\cline{3-7}
&&ADAM&51.52&63.64&63.64&63.64\\
\hline
\multirow{6}{*}{\textbf{\textit{Scan\_2}}}&\multirow{3}{*}{\textbf{$10^{-4}$}}&SGDM&60.61&57.58&57.58&57.58\\
\cline{3-7}
&&RMSprop&75.76&\done {\textbf{87.88*}}&81.82*&84.85\\
\cline{3-7}
&&ADAM&78.79&75.76&78.79&84.85\\
\cline{2-7}
&\multirow{3}{*}{\textbf{$10^{-3}$}}&SGDM&81.82&84.85&81.82*&81.82\\
\cline{3-7}
&&RMSprop&78.79&75.76*&81.82&69.70\\
\cline{3-7}
&&ADAM&66.67&72.73&72.73&66.67\\
\hline
\multirow{6}{*}{\textbf{\textit{Scan\_3}}}&\multirow{3}{*}{\textbf{$10^{-4}$}}&SGDM&72.73&75.76&66.67&69.70\\
\cline{3-7}
&&RMSprop&78.79*&72.73&84.85*&78.79\\
\cline{3-7}
&&ADAM&81.82*&69.70*&75.76&72.73*\\
\cline{2-7}
&\multirow{3}{*}{\textbf{$10^{-3}$}}&SGDM&78.79&75.76*&78.79*&72.73*\\
\cline{3-7}
&&RMSprop&75.76&78.79&72.73&75.76*\\
\cline{3-7}
&&ADAM&69.70&69.70&66.67&75.76\\
\hline
\end{tabular}
\end{adjustbox}\\
* validation patience 5
\end{table}

The analysis of table \ref{table4} also gives some interesting insights about the behaviour of model when the hyper-parameters change. These values were obtained after the 4-class classification of microscopy dataset. The parameters are most sensitive to scanning methods in this dataset as we could observe from the table \ref{table4} that when the patches extracted from \textit{Scan\_3} were trained using the same hyper-parameters, absolute drop in the accuracy was recorded. The results also indicate of the fact that scanning techniques can over-power the outcome of the model especially in the case of sequence modelling of images to labels. In Microscopy dataset as well, the learning rate $10^{-4}$ performed better than $10^{-3}$ and the scanning method \textit{Scan\_2} gave better outcome in comparison to other two methods. We observed the difference in optimization function (ADAM) and drop-out rate (0.6) when compared with best performing hyper-parameters in WSI dataset. 
The hyper-parameter tuning gave us the insight as  to how our model behaves which helped us to finally chose our parameter set to perform cross-validation. We deduced that learning rate $10^{-4}$ and scan technique \textit{Scan\_2} with validation patience gave us the better results in both the datasets. 
\begin{table}[htbp]
\renewcommand{\arraystretch}{1.6}
\caption{Accuracy (\%) obtained against different learning rates, drop-out rates, optimizing function and scanning techniques with respect to Microscopy Images (4-classes).}
\label{table4}
\centering
\begin{adjustbox}{width=3.5in}
\small
\begin{tabular}{|c|c|c|c|c|c|c|}
\hline
\multirow{2}{*}{\textbf{\shortstack{Scanning\\ Method}}}&\multirow{2}{*}{\textbf{\shortstack{Learning\\ Rate}}}&\multirow{2}{*}{\textbf{Optimizer}}&\multicolumn{4}{|c|}{\textbf{Drop-out Rate}}\\
\cline{4-7}
&&&\textbf{0.4}&\textbf{0.5}&\textbf{0.6}&\textbf{0.7}\\
\hline
\multirow{6}{*}{\textbf{\textit{Scan\_1}}}&\multirow{3}{*}{\textbf{$10^{-4}$}}&SGDM&59.32*&59.32&55.93&62.71\\
\cline{3-7}
&&RMSprop&74.58&72.88&74.58&69.49\\
\cline{3-7}
&&ADAM&76.27&72.88&79.66*&76.27\\
\cline{2-7}
&\multirow{3}{*}{\textbf{$10^{-3}$}}&SGDM&69.49&69.49&69.49&67.80\\
\cline{3-7}
&&RMSprop&67.80&69.49*&64.41*&62.71\\
\cline{3-7}
&&ADAM&59.32*&69.49*&59.32&71.19*\\
\hline
\multirow{6}{*}{\textbf{\textit{Scan\_2}}}&\multirow{3}{*}{\textbf{$10^{-4}$}}&SGDM&55.93&71.19*&72.88*&71.19*\\
\cline{3-7}
&&RMSprop&79.66&83.05*&77.97*&81.36\\
\cline{3-7}
&&ADAM&76.27&81.36&\done{\textbf{84.75*}}&76.27*\\
\cline{2-7}
&\multirow{3}{*}{\textbf{$10^{-3}$}}&SGDM&76.27&81.36*&83.05*&83.05*\\
\cline{3-7}
&&RMSprop&62.71&72.88*&79.66*&74.58*\\
\cline{3-7}
&&ADAM&71.19*&71.19*&81.36*&74.58*\\
\hline
\multirow{6}{*}{\textbf{\textit{Scan\_3}}}&\multirow{3}{*}{\textbf{$10^{-4}$}}&SGDM&18.69*&22.03*&22.03*&23.73*\\
\cline{3-7}
&&RMSprop&1.69*&1.69*&1.69*&1.69*\\
\cline{3-7}
&&ADAM&0&3.39*&0&3.39*\\
\cline{2-7}
&\multirow{3}{*}{\textbf{$10^{-3}$}}&SGDM&3.39*&5.08*&3.39*&5.08*\\
\cline{3-7}
&&RMSprop&10.17*&15.25*&11.86*&16.95*\\
\cline{3-7}
&&ADAM&8.47&15.25*&13.56*&22.03*\\
\hline
\end{tabular}
\end{adjustbox}\\
* validation patience 5
\end{table}
\par The direct analysis of comparative methods in literature \cite{ren2018differentiation,qaiser2019learning,
bychkov2018deep} with our proposed method could not be achieved since these methods have different objectives like calculating HER2 scores, five year disease specific survival prediction, hazard ratios. Also, they have different data values associated with each image to facilitate survival analysis on their datasets. Whereas, we do not have such type of data and hence the objectives are different. However, all these methods used CNN + LSTM as their backbone model. Therefore, for indirect qualitative analysis, we performed experiments with one LSTM layer instead of BilSTM layer while keeping all the other hyperparameters unchanged. For microscopy dataset, we achieved the 10 fold cross-validation overall accuracy of 88.75\%. Whereas, we achieved overall accuracy of 54.55\% for WSI dataset. Table \ref{lstmExp} records the classwise results with LSTM layer.
\begin{table*}[htbp]
\caption{Experimental results for proposed model with LSTM layer.}
\label{lstmExp}
\centering
\renewcommand{\arraystretch}{2.5}
\begin{adjustbox}{width=\textwidth}
\begin{tabular}{|c|c|c|c|c|c|c|c|c|c|}
\hline
\multicolumn{2}{|c|}{}&\multicolumn{2}{c}{\textbf{Benign}}&\multicolumn{2}{c}{\textbf{Invasive}}&\multicolumn{2}{c}{\textit{\textbf{In situ}}}&\multicolumn{2}{c|}{\textit{\textbf{Normal}}}\\
\hline
\textbf{Dataset}&\textbf{Acc}&\textbf{Se.}&\textbf{Sp.}&\textbf{Se.}&\textbf{Sp.}&\textbf{Se.}&\textbf{Sp.}&\textbf{Se.}&\textbf{Sp.}\\
\hline
\textbf{Microscopy}&0.8875$\pm$0.0056&0.8516$\pm$0.0106&0.9353$\pm$0.0050&0.9303$\pm$0.0103&0.9653$\pm$0.0061&0.8644$\pm$0.0081&0.9416$\pm$0.0039&0.9789$\pm$0.0045&0.9933$\pm$0.0014\\
\hline
\textbf{WSI}&0.5455$\pm$0.0629&0.4917$\pm$0.0410&0.8967$\pm$0.0116&0.7277$\pm$0.0115&0.5885$\pm$0.0421&0.7508$\pm$0.0481&0.8505$\pm$0.0376&-&-\\
\hline
\end{tabular}
\end{adjustbox}
\end{table*} 
From the obtained results, we observed that the performance with LSTM layer has degraded in comparison to BILSTM layer. Moreover, with WSI dataset, the performance degradation is quite significant relative to what we observe with Microscopy dataset. Therefore, besides philosophical justification, the short experimental observation also strengthened the choice of BiLSTM over LSTM layer. 

\subsection{\textbf{Results and comparison with current literature}}
The experiments on hyper-parameters tuning on both datasets gave us the optimal set to cross-validate the final accuracy value obtained over the two datasets. For benchmarking purpose we evaluated the performance of state of the art deep learning models- ResNet50\cite{he2016deep}, InceptionV3\cite{szegedy2017inception}, and  DenseNet201\cite{huang2017densely}. We used the constant learning rate of $10^{-4}$ and SGDM as optimizer for fine-tuning each of these networks. The last fully connected layer of each of these models were removed and replaced with our new fully connected layer having four outputs in the case of Microscopy dataset and three outputs for the WSI dataset. Each model was trained for 30 epochs. After the model training, we performed majority voting scheme to predict the final label for the image. This process was done for both Microscopy and WSI dataset. The benchmark models are not end-to-end due to the required post-processing of patch-based classifier outputs for image label prediction.  
\par We compared the results from benchmark models and the  results by top 5 teams in BACH grand challenge \cite{bach2018} published in \cite{aresta2019bach} with our proposed method on the Microscopy dataset in table \ref{comparisonMicro}. Similarly, for WSI dataset, we compared our model's performance in table \ref{comparisonWSI}. We performed 10 fold cross-validation on our proposed model. We have evaluated the performance of our model in terms of overall accuracy of the model, class-wise sensitivity, and specificity.  
\par Sensitivity and Specificity are commonly used for measuring medical applications. Sensitivity refers to how much our model is sensitive in detecting positive class or the percentage of actual positives that are correctly identified. Whereas, Specificity is the measure of actual negatives that are correctly identified. Both Sensitivity and Specificity of the model should be as high as possible to be able to correctly detect all positive samples and all negative samples. 
\begin{equation}
Sensitivity= \frac{True\:Positive}{True\:Positive + False\:Negative}
\end{equation} 
\begin{equation}
Specificity= \frac{True\:Negative}{True\:Negative + False\:Positive}
\end{equation} 
\begin{table*}[htbp]
\caption{Comparative Performance metrics with standard errors for patch-to-image classification model for Microscopy Dataset (4-classes)}
\label{comparisonMicro}
\centering
\renewcommand{\arraystretch}{3.0}
\begin{adjustbox}{width=\textwidth}
\begin{tabular}{|c|c|c|c|c|c|c|c|c|c|}
\hline
\multicolumn{2}{|c|}{}&\multicolumn{2}{c}{\textbf{Benign}}&\multicolumn{2}{c}{\textbf{Invasive}}&\multicolumn{2}{c}{\textit{\textbf{In situ}}}&\multicolumn{2}{c|}{\textbf{Normal}}\\
\hline
\textbf{Method}&\textbf{Acc}&\textbf{Se.}&\textbf{Sp.}&\textbf{Se.}&\textbf{Sp.}&\textbf{Se.}&\textbf{Sp.}&\textbf{Se.}&\textbf{Sp.}\\
\hline
\textbf{Ours (proposed)}&0.9000$\pm$0.0053&0.9400$\pm$0.0052&0.9467$\pm$0.0059&0.9800$\pm$0.0042&0.9800$\pm$0.0023&0.9600$\pm$0.0052&0.9500$\pm$0.0039&0.7200$\pm$0.0235&0.9500$\pm$0.0022\\
\hline
\textbf{\shortstack{Chennaswamy et al.,\\ 2018 (team 216)* \cite{aresta2019bach}}}&0.87&0.8&0.96&0.88&0.99&0.84&1.0&0.96&0.88\\

\textbf{\shortstack{Kwok et al.,\\ 2018 (team 248)* \cite{aresta2019bach}}}&0.87&0.72&0.96&0.92&0.96&0.88&0.97&0.96&0.93\\

\textbf{\shortstack{Brancati et al.,\\ 2018 (team 1)* \cite{aresta2019bach}}}&0.86&0.68&0.97&0.96&0.95&0.84&0.99&0.96&0.91\\

\textbf{\shortstack{Wang et al.,\\ 2018 (team 157)* \cite{aresta2019bach}}}&0.83&0.64&0.99&0.8&0.97&0.92&0.91&0.96&0.91\\

\textbf{\shortstack{Kone et al.,\\ 2018 (team 19)* \cite{aresta2019bach}}}&0.81&0.4&0.99&0.92&0.89&0.92&0.92&1.0&0.95\\

\textbf{\shortstack{Roy et al.,\\ 2019 * \cite{roy2019patch}}}&0.90&0.70&1.0&1.0&1.0&1.0&0.93&0.90&0.93\\
\hline
\textbf{ResNet50 \cite{he2016deep}}&0.8675$\pm$0.0083&0.8674$\pm$0.0239&0.9341$\pm$0.0052&0.9243$\pm$0.0121&0.9666$\pm$0.0048&0.8307$\pm$0.0090&0.9706$\pm$0.0047&0.8605$\pm$0.0113&0.9542$\pm$0.0103\\
\textbf{DenseNet201 \cite{huang2017densely}}&0.8900$\pm$0.0107&0.8624$\pm$0.0227&0.9697$\pm$0.0039&0.8408$\pm$0.0209&0.9836$\pm$0.0029&0.9170$\pm$0.0099&0.9440$\pm$0.0055&0.9416$\pm$0.0080&0.9575$\pm$0.0093\\
\textbf{InceptionV3 \cite{szegedy2015rethinking}}&0.8700$\pm$0.0066&0.8499$\pm$0.0163&0.9463$\pm$0.0049&0.8628$\pm$0.0116&0.9735$\pm$0.0045&0.8557$\pm$0.0122&0.9599$\pm$0.0035&0.9125$\pm$0.0145&0.9476$\pm$0.0094\\
\hline
\end{tabular}
\end{adjustbox}
\footnotesize *The standard error data for comparative literature is not available
\end{table*} 
\subsubsection{Performance on Microscopy dataset}
The accuracy of our model is 3\% more than the top performing team 216 \textit{Chennaswamy et al.}. The authors also used pre-trained CNNs instead of building their own custom model. They used ensemble of ResNet-101 \cite{he2016deep} and two DenseNet-161 \cite{huang2017densely} networks. In comparison to our model which is end-to-end, they first trained ResNet-101 and a DenseNet-161 using images normalized with breast-histology data
and then another DenseNet-161 with images fine-tuned with ImageNet normalization. During the testing, the majority voting scheme was used to declare the class of the input image from among the three classes predicted by the three models. Other notable difference between our model and theirs is that they used bilinear interpolation to  resize their image dimensions from $2048 \times 1536$ to $224 \times 224$ whereas, we did not use resized images since that would have decreased the quality of extracted features. We maintained the resolution and instead broke the image into patches to decrease the size of the input image. At the training time, for feature extraction step through GoogleNet, the patches were resized from $256 \times 256$ to $224 \times 224$. Second team on the leaderboard \textit{Kwok et al.} team 248 trained their model using images from both microscopy dataset and extracted patches from WSI dataset. Their 2-stage process first trained the ResNet-v2 \cite{szegedy2017inception} pre-trained on ImageNet on patches acquired from microscopy dataset and then again pre-trained their network with the patches acquired from WSI dataset. The prediction of each patch was then aggregated to image-wise prediction.  Their method was also not end-to-end and required two datasets to fine-tune the model performance. The difference between accuracy between our model and theirs was also 3\%. The class-wise comparison (Table \ref{comparisonMicro}) suggests that our model is much sensitive then the top 2 performing teams. Team 1 \textit{Brancati et al.} also used the ensemble of three ResNet models having 34, 50, and 101 layers, respectively. They used down-sampled microscopy images to extract patches of two sizes $308 \times 308$ and $615 \times 615$. These patches were taken from the center of the down-sampled images. They used highest class probability from three models as the class of the image. Our model performed better than theirs by overall accuracy of 90\% against 86\%. The next team in the list was Team 157 \textit{Wang et al.}. The authors in this work trained VGG16 \cite{simonyan2014very} using sample pairing data augmentation technique by \cite{inoue2018data} in which samples from different classes are augmented and then merged. The merged images are then trained using the chosen model. In the next step, the trained classifier from the mixed images is again trained using the initial non mixed dataset. The authors have resized their images to $256 \times 256$ and then extracted patches of size $224 \times 224$ at random locations. They achieved the accuracy of 83\%. The difference between their and our approach is same as with other competitive models. Team 19 \textit{Kone et al.} achieved the accuracy of 81\%, 9 percent less than our proposed model. They proposed binary tree like structure of 3 ResNeXt50 \cite{xie2017aggregated} models in which the top CNN in the hierarchy classifies images into carcinoma (\textit{In situ}, Invasive) and non-carcinoma{normal and benign}. The next two children of the root CNN then classifies the images into respective two sub-classes benign or normal and, \textit{In situ} or Invasive. They also used the two-stage process that used the learned weights of first stage to train the subsequent stages. All these methods in the challenge \cite{bach2018} who have reported their models performance used current state of the art deep learning models. The common thread between these models was that all used pre-trained models due to limited amount of data. However, they all used very heavy resizing of images which compromise with the quality of the high resolution intrinsic details present in cancer data. Moreover, their methods used two-three stages of training and the final outputs were aggregated to declare imagewise prediction. Our model on the other hand as mentioned avoid the disadvantages posed by the compared models. The same disadvantages are posed by the authors in \cite{roy2019patch} as well. They extracted different size patches ($64 \times 64, 128 \times 128, 512 \times 512$) to train their model separately but found optimum performance with $512 \times 512$. They then used heavy data augmentation to increase the amount of data. The augmented dataset is then trained using their custom CNN architecture. After the patches were trained, they used majority voting scheme to declare the predicted class of the input image. Although, they have achieved equal accuracy as our proposed model but suffered from the drawback of stage-wise model, data augmentation, and having to train their model from scratch which demands time and space. 
\begin{table*}[htbp]
\caption{Comparative Performance metrics with standard errors for patch-to-image classification model  for WSI Dataset (3-classes)}
\label{comparisonWSI}
\centering
\renewcommand{\arraystretch}{2.5}
\begin{adjustbox}{width=\textwidth}
\begin{tabular}{|c|c|c|c|c|c|c|c|}
\hline
\multicolumn{2}{|c|}{}&\multicolumn{2}{c}{\textbf{Benign}}&\multicolumn{2}{c}{\textbf{Invasive}}&\multicolumn{2}{c|}{\textit{\textbf{In situ}}}\\
\hline
\textbf{Method}&\textbf{Acc}&\textbf{Se.}&\textbf{Sp.}&\textbf{Se.}&\textbf{Sp.}&\textbf{Se.}&\textbf{Sp.}\\
\hline
\textbf{Ours (proposed)}&0.8402$\pm$0.0032&0.7090$\pm$0.0309&0.9132$\pm$0.0157&0.9142$\pm$0.0136&0.9190$\pm$0.0096&0.8333$\pm$0.0264&0.9240$\pm$0.0117\\
\hline
\textbf{ResNet50 \cite{he2016deep}}&0.8127$\pm$0.0093&0.9233$\pm$0.0202&0.8341$\pm$0.0148&0.8285$\pm$0.0113&0.9492$\pm$0.0126&0.7167$\pm$0.0271&0.9556$\pm$0.0065\\
\textbf{DenseNet201 \cite{huang2017densely}}&0.8127$\pm$0.0054&0.8142$\pm$0.0202&0.9091$\pm$0.0071&0.8520$\pm$0.0098&0.9160$\pm$0.0135&0.7833$\pm$0.0500&0.9056$\pm$0.0079\\
\textbf{InceptionV3 \cite{szegedy2015rethinking}}&0.8221$\pm$0.0087&0.8356$\pm$0.0140&0.8740$\pm$0.0142&0.8451$\pm$0.0077&0.9183$\pm$0.0087&0.7667$\pm$0.0245&0.9369$\pm$0.0696\\
\hline

\end{tabular}
\end{adjustbox}
\end{table*} 
\subsubsection{Performance on WSI dataset}
Microscopy dataset has balanced sets of four classes having equal image dimensions. The labelled mask of each class covers the entire image area and hence the features detected belong to one class only. These properties have helped to capture patches that completely belong to the labelled image class. however, with WSI dataset, due to arbitrary shape and size of the regions, the automatic extraction script could only extract the tumour from the surrounding bounding box area. Hence, the patches sampled from such WSI regions also contained a lot of non-tumour or non-class images. Moreover, the final acquired image regions were imbalanced (Table \ref{table2}). Therefore, these reasons might have caused the performance decline in the accuracy with WSI dataset in comparison to microscopy images. We trained for only three classes since the normal patches were randomly extracted and therefore, did not belong to one particular area in the WSI. The continuity of the patches is the important factor for our model. For experimental purposes when we trained our model with non-continuous normal patches, our model suffered from  performance decline which proved that the continuous patches draw spatial and contextual relationship through BiLSTMs. Otherwise, in the absence of non-continuity, the model may suffer from high variance. For benchmarking purposes and due to the lack of other comparative models, we compared our model with ResNet50, InceptionV3, And DenseNet121. From the Table \ref{comparisonWSI}, we could observe an improvement in the performance metrics when we used context based model. The main difference between our model and these state of the art models is that we did not train any deep architecture  and our model is end-to-end.

\section{Discussions}
Computer Aided Diagnosis (CAD) by analysing samples of Ultrasound, CT, and MRI images has been vastly suggested by medical image researchers for quite sometime. They trained machine learning models with various morphological, graph, and intensity based methods from very small set of data samples which were sometimes in the range of only 30 to 100 images.  The generalizing capability of such models has thus been questionable. However, after the introduction of deep learning models and availability of large amount of data. CAD techniques have experienced a huge success in performance precision and accuracy. When such deep models were tested for histopathological images, the low inter-class variability, especially between Normal and Benign classes, affected the overall performance. Hence, some new methods engaging these deep models in form of cascaded or ensemble architectures were proposed. Also, the most biopsy samples digitized at high resolutions contain very detailed information of cell structures and various other microstructures. The amount of information in one biopsy sample could collectively form a gigapixel image. Such high-resolution images are then required to be broken into smaller patches for further processing. Patch-based processing with complex ensemble methods followed by aggregation of patches into image labels in case of classification and segmented objects in case of segmentation makes it a lengthy process. The whole pipeline is divided into stages and lacks contextual relationship between patches. To overcome this drawback, we thought to streamline the process into an end-to-end network. The patches were visualized as a sequence of images as in a video and an effort was made to scan the patches so as to maintain as much continuity as possible. RNN based BiLSTM models are known to serve the purpose for predicting input sequence labels. Since, with BiLSTMs., we could capture both past and future contexts which enabled the model to aggregate the whole tumour features despite providing non-overlapping tumour parts in the form of patches as input sequence. \\
Due to sequence classification, the next step of  predicting image label from patch labels was not required. The graphical structure of BiLSTMs helped to build a context-based high-resolution tumour classification model that also gave us the benefit of end-to-end network structure. We also analysed that with our proposed models, there is no need of training deep models. We used a pre-trained ImageNet model for feature extraction and only one BiLSTM layer to train a shallow network. The average time to train the model was 17 minutes for 30 epochs. Once the model's hyper-parameters are tuned for the particular dataset, the training would take only few minutes. The shallow structure of the model also make it feasible for deployment in lighter applications such as hand-held devices like mobile handsets. The complexity of the method is discussed in \autoref{complexity}. Another advantage is that the various limitation of high-resolution images could be exploited in the favour of the methodology. The large dimensions could be easily turned into sequences using the appropriate scanning process. Due to BiLSTM layer, the model encapsulates the context mining capability which helped form the spatial and contextual relationship between patches sampled from a single image. The results suggested that this context modelling was crucial in patch-based models that process patches instead of complete structures at a time. In other words, modelling direct dependencies between patches, past or future, is crucial for performance of the model.
\par The idea of processing patches as a sequence using RNN based BiLSTM model could be further extended by using four RNNs. Each RNN would take patches going in up, down, left, and right directions, respectively \cite{bengio2017deep}. According to the \cite{visin2015renet, kalchbrenner2015grid}, compared to CNNs, RNNs when applied to images allow for long-range lateral interactions between features in the same feature map. 
\section{Complexity}
\label{complexity}
The model is end-to-end deep learning model whose architecture is briefly expressed in \autoref{network}. Till layer number $143- Flatten Layer$, there are no Floating Point Operations (FLOPs) being performed. GoogLeNet network is present to extract pre-trained features which are then passed on to subsequent layers for further processing. Similarly, Sequence Folding, Unfolding, Average Pooling, and Flatten layer also accumulate zero FLOPs. Therefore, the time complexity is calculated from BiLSTM layer onwards. The formula for calculating number of learnable parameters in a BiLSTM layer is derived as follows, \\
Let $I$ be the input size of the sequence, $K$ be the number of output dimensions and $H$ be the number of hidden units.
For BiLSTM if $H$ are the number of initialized hidden units then $M=2\times H$ are the total number of hidden units for both forward and backward passes of the BiLSTM network.
After concatenation of the forward and backward outputs, the total output dimensions become $K=M/2$.
Then the complexity of a BiLSTM layer is:
\[\mathcal{O}(W)\]
where $W$ are the total number of learnable parameters in the network calculated as:
\[W=4\times M((I+1)+K)\] 
\[W=4\times (M(I+1)+MK)\]
Here in the above formula, the first term $4\times M(I+1)$ are the total number of input weights and the second term $4\times MK$ are the number of recurrent weights.\\
In the terms of Big Oh notation, the time complexity is;
\[\mathcal{O}(M(I+1)+MK)\]
The multiplication by factor $4$ represents four weight matrices of BiLSTM layer (Input gate, Forget gate, Cell candidate, Output gate). The input size variable $I$ is added with a bias value $1$. \\
For the BiLSTM layer in our network, the number of parameters are:
\[W=4\times4000\times ((1024+1)+2000)\]
\[W=16000 \times (1025+2000)\]
\[W=48400000\]
where $4000$ are the total number of hidden units for both forward and backward passes of the BiLSTM layer, $1024$ is the size of the input sequence, and $2000$ is the total number of outputs.\\
Next, for the fully connected layer, the parameters are 
\[F=3\times 4000\]
Hence, the total number of FLOPs are \[W+F=48400000+12000=48412000\]
\[W+F=48.4 \times 10^6 = 48 MFLOPs\]

We have used NVIDIA TitanX GPU (12GB) for training our models. It performs $11 \times 10^{12}$ or 11 Tera FLOPs per second which is a sufficient computational efficiency required for training. \\
To put it in perspective, we mention the number of FLOPs for few popular deep learning networks in Table \ref{FLOPs}. \\
\begin{table}[htbp]
\renewcommand{\arraystretch}{1.6}
\caption{FLOPs for popular deep learning architectures}
\label{FLOPs}
\centering
\begin{tabular}{|c|c|}
\hline
AlexNet&727 MFLOPs\\
VGG16&16 GFLOPs\\
VGG19&20 GFLOPs\\
GoogLeNeT&2 GFLOPs\\
ResNet50&4 GFLOPs\\
DenseNet121&3 GFLOPs\\
InceptionV3&6 GFLOPs\\
\hline
\end{tabular}
\end{table}
In terms of the Big Oh notation, the time complexity of the model for $t$ number of input samples and $n$ number of epochs is represented as;
\[\mathcal{O}(n\times t\times (W+F))\]
\section{Conclusion}
We proposed an end-to-end RNN based model that takes patches as input and outputs image labels. The patches are modelled as sequences by using one-layer BiLSTM model. The sequence in an image is captured using the strategic scanning method which was experimentally chosen. We used BACH challenge dataset to test our method and reported our results on two different datasets introduced in the challenge. The classifier performance was compared with recently reported metrics by top 5 teams in BACH challenge for microscopy dataset. We achieved highest performance of 90\% with simpler architecture and less time and space complexity.

\section*{Acknowledgement}
This research was carried out in Indian Institute of Information Technology, Allahabad and supported, in part, by the Ministry of Human Resource and Development, Government of India and the Biomedical Research Council of the Agency for Science, Technology, and Research, Singapore. We are also grateful to the NVIDIA corporation for supporting our research in this area by granting us TitanX (PASCAL) GPU.

\bibliographystyle{elsarticle-num}
\bibliography{classificationTumours}

\end{document}